\documentclass[sigconf]{acmart}
\usepackage{multirow}
\usepackage{balance}

\AtBeginDocument{%
  }

\AtBeginDocument{%
}

\copyrightyear{2026}
\acmYear{2026}
\setcopyright{cc}
\setcctype{by}
\acmConference[UbiComp Companion '26]{Companion of the 2026 ACM International Joint Conference on Pervasive and Ubiquitous Computing}{October 11--15, 2026}{Shanghai, China}
\acmBooktitle{Companion of the 2026 ACM International Joint Conference on Pervasive and Ubiquitous Computing (UbiComp Companion '26), October 11--15, 2026, Shanghai, China}
\acmDOI{10.1145/3798063.3839146}
\acmISBN{979-8-4007-2533-3/2026/10}

\begin{document}

\title{Self-Supervised Pretraining for Label-Efficient Facial Expression
  Recognition using Smart Glasses}

\author{Zhengtao Gao}
\orcid{0009-0008-9873-143X}
\email{zhengtao.gao@usi.ch}
\affiliation{%
  \institution{Università della Svizzera italiana}
  \city{Lugano}
  \country{Switzerland}
}

\author{Francesco Bombassei De Bona}
\orcid{0000-0003-2120-4187}
\email{francesco.bombassei.de.bona@usi.ch}
\affiliation{%
  \institution{Università della Svizzera italiana}
  \city{Lugano}
  \country{Switzerland}
}

\author{Leonardo Alchieri}
\email{leonardo.alchieri@usi.ch}
\orcid{0000-0003-3088-6132}
\affiliation{%
  \institution{Università della Svizzera italiana}
  \city{Lugano}
  \country{Switzerland}
}

\author{Ivana Kiprijanovska}
\orcid{0000-0002-1700-0344}
\email{ivana.kiprijanovska@emteqlabs.com}
\affiliation{%
  \institution{Emteq Ltd.}
  \city{Brighton}
  \country{United Kingdom}
}

\author{Simon Stankoski}
\orcid{0000-0002-0377-6731}
\email{simon.stankoski@emteqlabs.com}
\affiliation{%
  \institution{Emteq Ltd.}
  \city{Brighton}
  \country{United Kingdom}
}

\author{Bojan Sofronievski}
\orcid{0009-0003-9701-3229}
\email{bojan.sofronievski@emteqlabs.com}
\affiliation{%
  \institution{Emteq Ltd.}
  \city{Brighton}
  \country{United Kingdom}
}

\author{Hristijan Gjoreski}
\orcid{0000-0002-0770-4268}
\affiliation{%
  \institution{Ss. Cyril and Methodius University in Skopje}
  \city{Skopje}
  \country{North Macedonia}
}
\additionalaffiliation{%
  \institution{Emteq Ltd.}
  \city{Brighton}
  \country{United Kingdom}
}
\email{hristijang@feit.ukim.edu.mk}

\author{Marc Langheinrich}
\orcid{0000-0002-8834-7388}
\affiliation{%
  \institution{Università della Svizzera italiana}
  \city{Lugano}
  \country{Switzerland}
}
\email{marc.langheinrich@usi.ch}

\author{Martin Gjoreski}
\orcid{0000-0002-1220-7418}
\affiliation{%
  \institution{Università della Svizzera italiana}
  \city{Lugano}
  \country{Switzerland}
}
\email{martin.gjoreski@usi.ch}

\renewcommand{\shortauthors}{Zhengtao Gao et al.}

\begin{abstract}
  Facial Expression Recognition (FER) on optomyographic (OMG) smart glasses such as OCOsense enables privacy-preserving, camera-free affect sensing for health and well-being. Most FER systems are fully supervised, i.e., they rely either on hand-crafted expert features with classical machine learning, or on end-to-end deep models trained from labelled data. Both depend on large quantities of labelled expressions, which are costly to collect and limit deployment in real-world, personalized settings. We investigate Self-Supervised Learning (SSL) as a route to label-efficient FER on these glasses.
  We pretrain a Vision Transformer with a masked-autoencoder objective on unlabelled OCOsense recordings and finetune it for five-class FER under person-independent cross-validation. Across label budgets from 5\% to 100\%, pretraining consistently improves performance; with only 5\% of the labels, the pretrained model outperforms the from-scratch model trained with all labels.
  Our results show that SSL pretraining is a practical way to reduce labelling cost for smart eyewear FER, and we discuss the implications for future personalization and continual adaptation.
\end{abstract}



\maketitle

\section{Introduction}
Mental-health and well-being disorders are among the leading causes of
disability worldwide, affecting roughly 16--20\% of the population in many
regions, yet most individuals rarely receive timely or personalized
support~\cite{arias2022quantifying}. Wearable devices now make continuous,
in-the-wild behavioral and physiological monitoring feasible at scale, opening
the door to earlier and more personalized well-being sensing
~\cite{broulidakis2023optomyography}. A persistent obstacle, however, is robust
generalization across users, devices, and real-world conditions. Behavioral and
affective signals vary substantially from person to person, and models that
perform well on a training cohort often degrade on new, unseen
users~\cite{han2024systematic}. Methods that adapt to individuals without
demanding large per-user labelled datasets are therefore key to enabling
practical well-being sensing with smart devices.

Smart eyewear is becoming a practical platform for continuous, context-aware
health and well-being sensing. Because the frame sits stably on the face and
head, it can host sensors that observe facial-muscle activity and head motion
without a camera pointed at the wearer, sidestepping the positioning and
privacy problems that limit camera-based facial analysis on wearables.
Optomyographic (OMG) glasses such as OCOsense~\cite{archer2023ocosense} use
non-contact optical sensors to measure skin movement over major facial muscles,
and have been shown to support facial expression recognition (FER) for affect
and well-being monitoring~\cite{kiprijanovska2023towards, marinova2025deep}.

Despite these advances, FER systems for smart glasses---across camera
~\cite{sajjad2023comprehensive}, Electromyography
(EMG)~\cite{cha2020realtime,gjoreski2022facial}, Electrooculography (EOG)
~\cite{rostaminia2019w}, acoustic ~\cite{li2024eyeecho, xie2021acoustic},
capacitive ~\cite{latino2026fer}, and OMG
sensing~\cite{kiprijanovska2023towards, marinova2025deep}---are predominantly
fully supervised. They typically extract hand-crafted features for classical
machine learning classifiers or train deep networks end-to-end, in both cases
depending on labelled expression data. Because facial and expression morphology
varies across users, models trained on one cohort transfer imperfectly to new,
unseen users---the person-independent domain-shift problem in wearable
sensing~\cite{han2024systematic}. Collecting large labelled datasets per
deployment is impractical, motivating methods that learn useful representations
from unlabelled data. This limitation is shared by the OMG-based OCOsense
systems on which we perform our experiments~\cite{archer2023ocosense,
  kiprijanovska2023towards,marinova2025deep}.

We study Self-Supervised Learning (SSL) as a way to reduce the labelling burden
for FER. SSL aims to learn transferable representations from large amounts of
unlabelled data through automatically generated supervisory signals. We
pretrain a Vision-Transformer (ViT) encoder on unlabelled smart-glass data with
a masked-autoencoder (MAE) objective, then finetune it for FER under a
person-independent cross-validation. Our primary research question is:
\textit{can self-supervised pretraining improve OMG-based FER in label-scarce
  settings?}

Our contributions are the following: (i) We present, to our knowledge, the
first self-supervised pretraining approach for FER on OMG smart glasses; and
(ii) we characterize label-efficiency by varying the labelled-data budget,
while comparing our SSL-pretrained approach against supervised training from
scratch.

\section{Related Work}\label{sec:related_work}

\paragraph{FER using smart glasses}

Smart-eyewear facial sensing has been explored using cameras, EMG, EOG, active
acoustics, and capacitive
sensors~\cite{sajjad2023comprehensive,cha2020realtime,gjoreski2022facial,rostaminia2019w,li2024eyeecho,xie2021acoustic,latino2026fer}.
These modalities offer different trade-offs in privacy, skin contact,
robustness, and sensing range. However, none of these studies investigates SSL
pretraining for label-efficient FER.

Researchers have explored the use of OMG-based OCOsense glasses for
FER~\cite{kiprijanovska2023towards}. Their feature-based approach extracts
statistical descriptors from OMG sensor data and trains a Random Forest
classifier to recognize four posed expressions and a neutral state. Using
leave-one-subject-out cross-validation on data from 27 participants, the
authors report 93\% accuracy and a macro F1 score of 0.90. They further show
that head movement does not significantly degrade sensing performance, whereas
glasses placement has a noticeable effect. A subsequent
study~\cite{marinova2025deep} explored end-to-end deep learning for FER using
OCOsense glasses. The authors considered six classes: the same five with an
additional ``other'' category including eating, speaking, and bruxism (teeth
grinding). They compared seven convolutional architectures together with the
ConvBoost augmentation framework~\cite{shao2023convboost} on data from 55
participants. Their best performing model, Attention ConvLSTM with ConvBoost,
achieved a macro F1 score of 0.78. The authors highlight that the inclusion of
the ``other'' class and the larger cohort substantially increase task
difficulty compared with the original five-class posed setting.

\paragraph{SSL for wearable sensing}
Researchers have recently explored the use of SSL for wearable
sensing~\cite{logacjov2024self}, e.g. masked-autoencoder pretraining
(MAE)~\cite{he2022masked} on unlabelled wearable
data~\cite{narayanswamy2025scaling}, and a growing line of work builds
foundation models over large heterogeneous physiological data, from wearables
to modality-specific encoders such as EMG~\cite{narayanswamy2025scaling,
  khasentino2025personal, delpup2023applications, fasulo2025tinymyo}.

These studies show that SSL representations transfer across downstream tasks
and can reduce labelled-data requirements, including for affect and emotion
recognition from wearable physiological signals~\cite{dissanayake2022sigrep,
  monteroquispe2022applying}. However, most existing efforts focus on wrist- and
body-worn signals, while, to the best of our knowledge, no work has explored
SSL pretraining for facial sensing with on-glasses physiological sensors.

\section{Device and Data}

\begin{figure}[t]
  \centering
  \includegraphics[width=0.8\columnwidth]{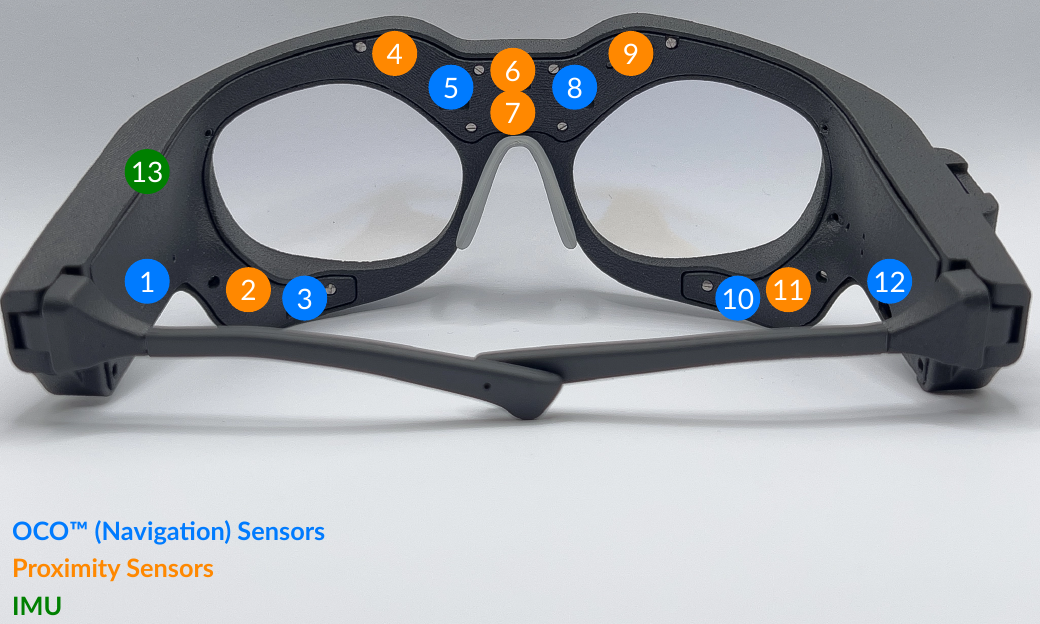}
  \caption{OCOsense OMG smart glasses and their sensors.}
  \Description{Photograph-style diagram of the OCOsense smart-glasses frame, annotated with the positions of the non-contact optical sensors over the frontalis, zygomaticus, corrugator, orbicularis and temple regions, together with the location of the 9-axis inertial measurement unit.}
  \label{fig:glasses}
\end{figure}

\paragraph{OCOsense smart glasses} We use the OCOsense smart glasses~\cite{archer2023ocosense}.
\autoref{fig:glasses} illustrates the OCOsense glasses and the placement of
the facial-sensing sensors on the glasses frame. The device carries non-contact
optical sensors that measure skin movement over the frontalis, zygomaticus,
corrugator, orbicularis, and temple regions, as well as a 9-axis IMU. All
sensors have a sampling rate of 50~Hz, with the OMG and proximity sensors
operating 4 to 30~mm above the skin. For FER, we exclude the magnetometer and
use a total of 30 sensor channels, comprising 6 IMU channels (3-axis
accelerometer and 3-axis gyroscope), 6 Euler-angle channels (sine and cosine
representations of roll, pitch, and yaw), 12 navigation channels (x/y
displacement at the left/right cheek, brow, and temple), and 6 proximity
channels corresponding to the six sensing locations. These signals form the
multichannel windows used throughout our pretraining and FER experiments.

\paragraph{Signal preprocessing}
Each stream is filtered, standardized per channel using session-level
statistics, and segmented into fixed windows (50\% overlap for MAE pretraining,
90\% for FER, with majority-vote labels).\footnote{Zero-phase third-order
  Butterworth filters. IMU: median filter (kernel~3) + 10~Hz low-pass, with a
  0.3~Hz component subtracted from the accelerometer to remove gravity.
  Navigation: cumulative sum + 0.1--5~Hz band-pass; proximity: same band-pass.
  Euler angles: median + 10~Hz low-pass, 0.3~Hz baseline removed from yaw,
  encoded as sine/cosine.} At 50~Hz the pretraining and FER windows are the
$(500,30)$ and $(50,30)$ encoder inputs of \autoref{sec:method}.

\paragraph{Pretraining dataset: CLUES}
We pretrain on unlabelled OCOsense recordings from the CLUES dataset, collected
from 52 participants across two sites~\cite{Krstevska2026CLUES}. The CLUES
dataset contains OCOsense recordings in a lab-controlled environment, where
users performed cognitive-load tasks alternating with relaxed breaks, eliciting
naturally occurring facial and head movements. After preprocessing, CLUES
yielded 29,722 overlapping 10~s windows, corresponding to 82.6 hours of model
input.

\paragraph{Downstream FER dataset: OCOGEM}
For the supervised FER task we used an extended version of the OCOGEM dataset
\cite{marinova2025deep}, in which 75 participants performed deliberately posed
facial expressions (smile, frown, eyebrow raise, and squeezed eyes) while
wearing the OCOsense glasses, with a neutral configuration maintained between
poses. Expressions were always performed in a fixed order, cued by an
instruction video with visual (on-screen emoticons) and auditory (beep)
prompts. This protocol yields a labelled dataset spanning variation in
expression intensity, duration (1--3~seconds), and head movement. The
pretraining corpus and the FER cohort do not share participants.

\begin{figure*}[t]
  \centering
  \includegraphics[width=\textwidth]{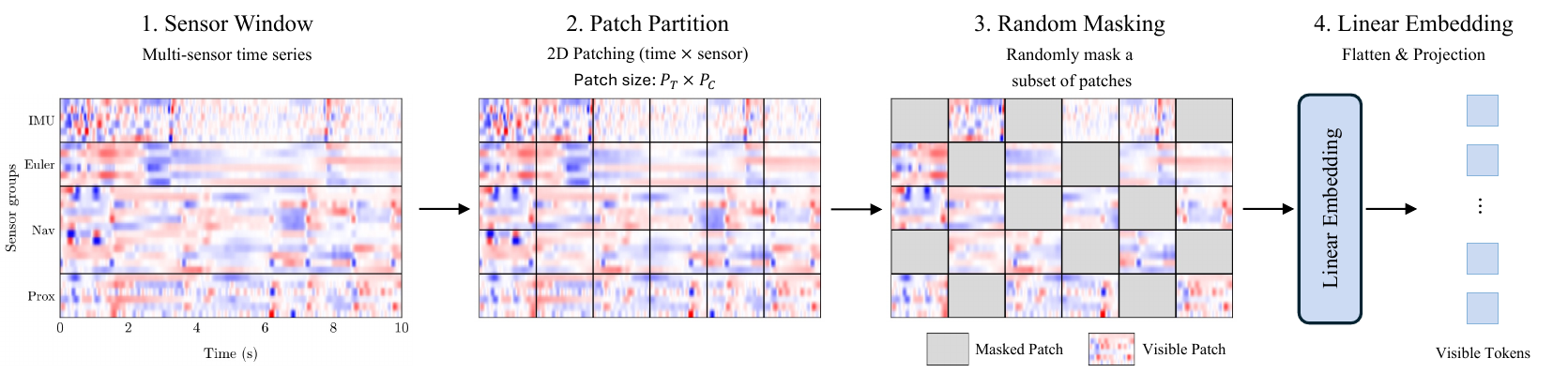}
  \caption{Illustration of the sensor arrangement and tokenization process. Multichannel sensor windows are partitioned into non-overlapping patches along the temporal and channel dimensions and projected into patch tokens for the ViT encoder.}
  \Description{Schematic of the tokenization scheme. A multichannel sensor window is drawn as a two-dimensional grid with time along the horizontal axis and the 30 sensor channels along the vertical axis, grouped by modality into IMU, Euler-angle, navigation and proximity blocks. The grid is divided into non-overlapping rectangular patches of 5 time steps by 6 channels, so that patch boundaries coincide with modality boundaries and each patch spans a single modality. Arrows show the patches being flattened and passed through a shared linear projection into a sequence of patch tokens that is fed to the Vision Transformer encoder.}
  \label{fig:arch}
\end{figure*}

\section{Method}\label{sec:method}

Our pipeline consists of MAE pretraining followed by supervised FER finetuning.
We also consider a supervised from-scratch baseline that uses the same encoder
architecture but is trained from scratch on the FER task without MAE
pretraining.

\paragraph{Tokenization and encoder}
Each input sample is a multichannel sensor window from the smart glasses. Along
the channel dimension, channels are grouped by sensor modality: IMU,
Euler-angle, navigation-sensor, and proximity-sensor channels. The input sensor
window is then partitioned into non-overlapping two-dimensional patches along
the time and sensor dimensions. The patch size is $(5,6)$ along the temporal
and channel dimensions, respectively, so channel boundaries align with modality
groups. Consequently, each patch contains channels from only a single sensor
modality and can be interpreted as a short temporal segment from that modality.
All sensor patches are then projected into token vectors using a shared linear
projection layer. The tokenization process is illustrated in
\autoref{fig:arch}.

The resulting token sequence is processed by a ViT
encoder~\cite{dosovitskiy2021image}. To preserve the two-dimensional structure
of the original window, we apply 2D RoPE~\cite{su2024roformer} in the
transformer blocks. We also prepend four learnable register
tokens~\cite{darcet2024vision} to the patch-token sequence, allowing the model
to store and propagate global information that might otherwise be forced into
ordinary patch tokens.

\paragraph{Masked-autoencoder pretraining}
We use MAE for pretraining~\cite{he2022masked}: patch tokens are randomly
masked independently for each sample. Only the visible tokens, together with
the register tokens, are passed through the encoder. The encoded visible tokens
are projected into a decoder space, concatenated with learned mask tokens,
restored to the original patch order, and processed by a lightweight decoder
that reconstructs the original patch contents.

The reconstruction target is the original sensor window, and the loss is
computed only over masked patches as:
\begin{equation}
  \mathcal{L}_{\mathrm{MAE}}
  = \frac{\sum_i m_i \cdot \mathrm{MSE}(\hat{p}_i, p_i)}{\sum_i m_i},
  \qquad m_i \in \{0,1\},
\end{equation}
where $p_i$ is the original patch, $\hat{p}_i$ is the reconstructed patch, and $m_i=1$ indicates that patch $i$ was masked.

\paragraph{Data augmentation}
During pretraining we apply four augmentations to the encoder input while
keeping the reconstruction target as the clean signal, so the model learns
representations robust to deployment nuisances. Each augmentation is applied
independently with probability $p=0.5$. \emph{Channel switching} exploits the
bilateral symmetry of the navigation-sensor layout: we exchange the paired
left/right navigation channels and flip the sign of their lateral ($x$-axis)
coordinates, modelling side-dependent sensor discrepancies. \emph{Jitter} adds
i.i.d. Gaussian noise ($\sigma=0.03$) to the IMU channels to mimic measurement
noise. \emph{Scaling} multiplies each channel by a per-window factor drawn from
$\mathcal{N}(1,0.02^2)$, exposing the model to cross-session amplitude
differences that survive session-level normalization. \emph{Magnitude warping}
multiplies each channel by a smooth random curve (interpolated from a few
control points, magnitude $0.02$) that varies slowly around one, approximating
slow gain drift over the window.

\paragraph{Finetuning for FER}
After pretraining, we discard the MAE decoder and attach a linear
classification head to the pretrained ViT encoder. The window representation is
obtained by global average pooling over the encoder's patch-token outputs.
During finetuning, the entire model is optimized end-to-end for the five-class
FER.

\paragraph{Evaluation protocol}
We evaluate our experiments using a person-independent five-fold
cross-validation, i.e., data from each participant is present inside a single
fold only. In each cross-validation run, one fold is used as the test set. From
the remaining four folds, 20\% of the participants are held out for validation,
and the rest are used for training. Validation data are used for checkpoint
selection.

\paragraph{Model Implementation Details}

The ViT encoder has an embedding dimension of 128, six Transformer blocks, four
attention heads, four register tokens, and 2D RoPE. The decoder uses an
embedding dimension of 64, three Transformer blocks, and four attention heads.
MAE pretraining and FER training use the $(500,30)$ and $(50,30)$ input
windows, respectively, both with a patch size of $(5,6)$. All models are
optimized using AdamW optimizer~\cite{adamw} with a batch size of 128. For MAE
pretraining, we use a learning rate of $3\times10^{-4}$, betas $(0.9,0.95)$,
weight decay $10^{-4}$, gradient clipping of 1.0, and a masking ratio of 75\%.
For FER training, we use betas $(0.9,0.99)$, weight decay $10^{-4}$, gradient
clipping of 1.0, label smoothing of 0.05, and head dropout of 0.5. The learning
rates are $4\times10^{-4}$ for training the from-scratch baseline and
$2\times10^{-5}$ for fine-tuning the MAE-pretrained model, respectively. All
models use a learning-rate schedule consisting of 5\% linear warmup followed by
cosine decay. MAE pretraining is performed for 100 epochs. To ensure comparable
optimization across the varying training-set sizes, models in the
label-efficiency experiments are trained for a fixed budget of 22,200
optimization steps and validated every 3,700 steps. Sensor-modality ablations
use the complete training set and run for six epochs per fold, with validation
after every epoch. In both settings, the checkpoint maximizing the sum of
validation macro F1 and AUC is used to evaluate on the test set.

\section{Experiments and Results}

\subsection{Label availability analysis}

\begin{figure}[t]
  \centering
  \includegraphics[width=0.9\columnwidth]{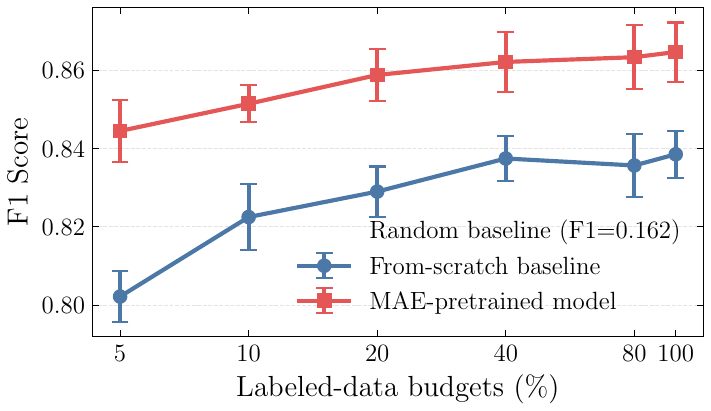}
  \caption{FER data efficiency comparison between MAE-pretrained model and from-scratch baseline.}
  \Description{Line plot of macro F1 score on the vertical axis against the labelled-data
    budget on the horizontal axis, at 5, 10, 20, 40, 80 and 100 percent of the labels. Two
    lines with error bars at each point are shown: the MAE-pretrained model and the
    from-scratch baseline. Both curves rise steeply from the 5 percent budget and then
    flatten out, and the MAE-pretrained curve lies above the from-scratch curve at every
    budget, with the gap widest at 5 percent labels. The MAE-pretrained model at the 5
    percent budget already reaches a higher F1 score than the from-scratch baseline trained
    with all of the labels.}
  \label{fig:fer_eff_f1}
\end{figure}

\paragraph{Overall results}
To probe label efficiency, we subsample the downstream training data within
each person-independent fold to 5\%, 10\%, 20\%, 40\%, 80\% and 100\%. Sampling
is performed within each user-label group so that the reduced training sets
remain balanced across participants and expression classes; validation and test
sets are left intact.

At each labelled-data budget, we compare the MAE-pretrained model with the
supervised from-scratch baseline. We also report the results from a random
baseline, which samples uniformly the five expression classes (averaged across
five seeds).

\autoref{fig:fer_eff_f1} shows the FER F1-score across labelled-data budgets, and \autoref{tab:labeleff} reports the corresponding accuracy, macro F1, AUC, and MCC.
The MAE-pretrained model consistently outperforms the supervised from-scratch baseline across all metrics and budgets, with the largest gain at the 5\% budget.

\begin{table*}
  \centering
  \caption{FER data efficiency comparison between the from-scratch baseline and the MAE-pretrained model.}
  \label{tab:labeleff}
  \begin{tabular}{ccccccccc}
    \toprule
    \multirow{2}{*}{Data}   &
    \multicolumn{2}{c}{Acc} &
    \multicolumn{2}{c}{F1}  &
    \multicolumn{2}{c}{AUC} &
    \multicolumn{2}{c}{MCC}                                                                                                                                                                 \\
    \cmidrule(lr){2-3}
    \cmidrule(lr){4-5}
    \cmidrule(lr){6-7}
    \cmidrule(lr){8-9}
                            & FS                & Pretrain          & FS                & Pretrain          & FS                & Pretrain          & FS                & Pretrain          \\
    \midrule
    5\%                     & $0.868 \pm 0.004$ & $0.898 \pm 0.006$ & $0.802 \pm 0.006$ & $0.845 \pm 0.008$ & $0.945 \pm 0.006$ & $0.977 \pm 0.003$ & $0.768 \pm 0.009$ & $0.818 \pm 0.010$ \\
    10\%                    & $0.883 \pm 0.004$ & $0.903 \pm 0.004$ & $0.823 \pm 0.008$ & $0.851 \pm 0.005$ & $0.968 \pm 0.002$ & $0.978 \pm 0.002$ & $0.792 \pm 0.009$ & $0.827 \pm 0.006$ \\
    20\%                    & $0.887 \pm 0.005$ & $0.908 \pm 0.005$ & $0.829 \pm 0.006$ & $0.859 \pm 0.007$ & $0.971 \pm 0.002$ & $0.980 \pm 0.002$ & $0.799 \pm 0.008$ & $0.836 \pm 0.008$ \\
    40\%                    & $0.892 \pm 0.004$ & $0.910 \pm 0.005$ & $0.837 \pm 0.006$ & $0.862 \pm 0.008$ & $0.975 \pm 0.003$ & $0.980 \pm 0.001$ & $0.810 \pm 0.008$ & $0.840 \pm 0.009$ \\
    80\%                    & $0.893 \pm 0.005$ & $0.911 \pm 0.006$ & $0.836 \pm 0.008$ & $0.863 \pm 0.008$ & $0.973 \pm 0.003$ & $0.980 \pm 0.002$ & $0.808 \pm 0.010$ & $0.841 \pm 0.010$ \\
    100\%                   & $0.894 \pm 0.004$ & $0.912 \pm 0.005$ & $0.839 \pm 0.006$ & $0.865 \pm 0.008$ & $0.974 \pm 0.003$ & $0.981 \pm 0.002$ & $0.810 \pm 0.006$ & $0.843 \pm 0.009$ \\
    \bottomrule
  \end{tabular}
\end{table*}

\begin{figure}[t]
  \centering
  \includegraphics[width=\columnwidth]{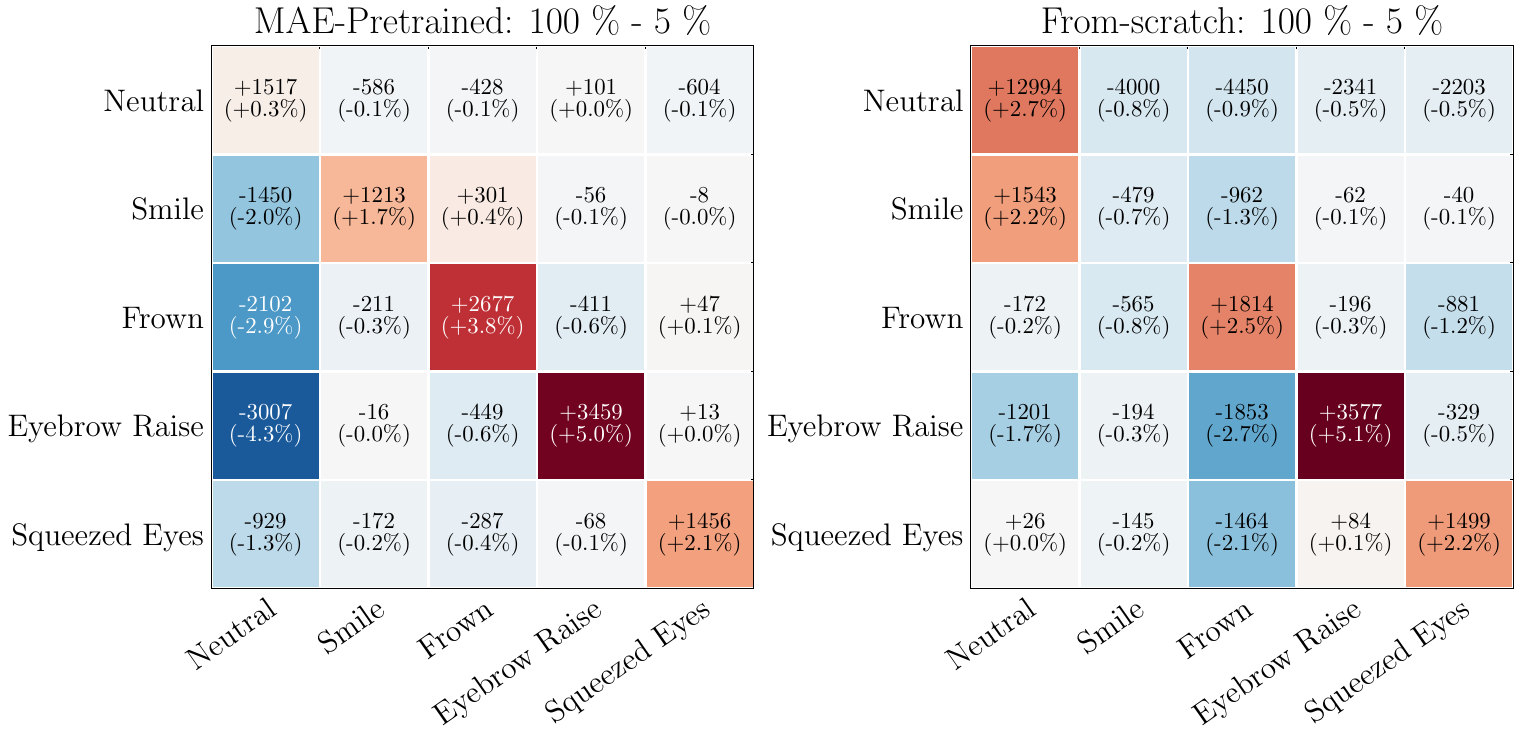}
  \caption{Data-scaling effect on class-level predictions. The matrices show
    100\%-minus-5\% confusion differences for the pretrained model and the
    from-scratch model. Rows denote ground-truth labels and columns denote predicted labels.}
  \Description{Two five-by-five heatmaps shown side by side, one for the MAE-pretrained
    model and one for the from-scratch model. Rows are the ground-truth classes and columns
    the predicted classes, in the order Neutral, Smile, Frown, Eyebrow Raise and Squeezed
    Eyes. Each cell holds the difference between the row-normalized confusion matrix at the
    100 percent label budget and the one at the 5 percent budget, encoded with a diverging
    colour scale centred at zero, where positive values indicate that the larger budget sends
    more probability mass to that cell. In both panels the diagonal cells are predominantly
    positive and the off-diagonal cells predominantly negative, with the largest diagonal
    gains for Eyebrow Raise and Frown. For the pretrained model the negative off-diagonal
    cells are concentrated in the Neutral column, whereas the from-scratch panel additionally
    shows large changes in the Neutral row and mutual confusions among Frown, Eyebrow Raise
    and Squeezed Eyes.}
  \label{fig:confusion}
\end{figure}

\paragraph{Label availability analysis on individual facial expressions}
We next analyze how label availability affects the different facial expression
classes. For each model and label budget, we sum the confusion-matrix counts
across the five participant-disjoint test folds. Because these test folds
jointly cover the complete dataset, each sample contributes exactly once to the
aggregated confusion matrix. After row normalization, \autoref{fig:confusion}
reports the element-wise difference $\mathrm{CM}_{100\%}-\mathrm{CM}_{5\%}$ for
the MAE-pretrained model and the from-scratch baseline. For both models, the
larger budget primarily increases the diagonal entries while reducing
off-diagonal confusions, indicating improved class separation. The one
exception is Smile for the from-scratch model, whose diagonal entry decreases
slightly, by 0.7 percentage points, from the 5\% to the 100\% budget. The
largest improvements are observed for the more expressive classes, particularly
Eyebrow Raise and Frown. For the MAE-pretrained model, additional labels mainly
reduce expressive samples being misclassified as Neutral, whereas the
from-scratch model exhibits larger changes in the Neutral class as well as
confusions among Frown, Eyebrow Raise, and Squeezed Eyes. The relatively
smaller improvement for the Neutral class in the pretrained model suggests that
the 5\%-label model already learns a reliable representation for neutral
expressions, allowing the additional labels to focus primarily on refining the
more challenging expressive classes.

\begin{figure}[t]
  \centering
  \includegraphics[width=\columnwidth]{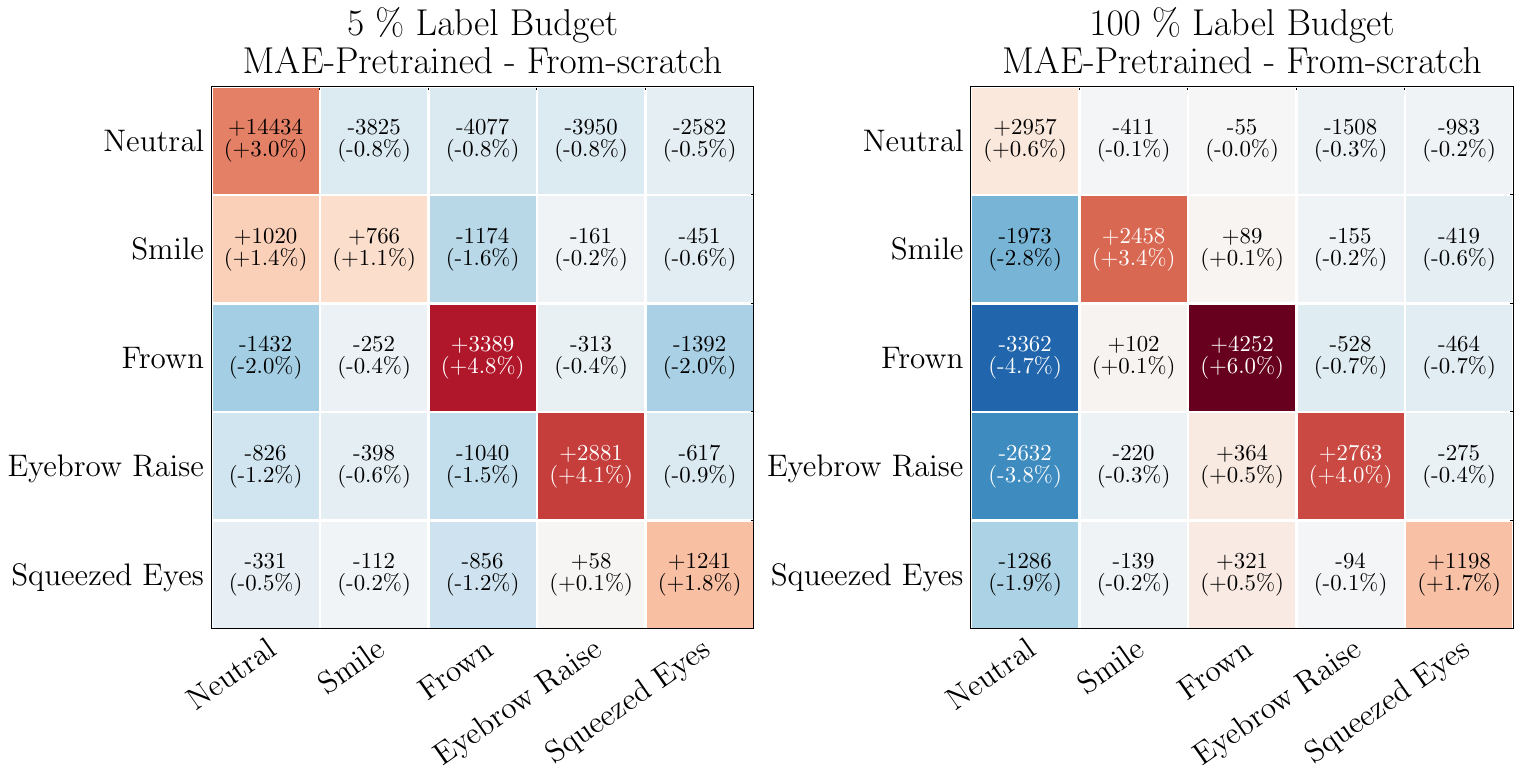}
  \caption{Class-level effect of MAE pretraining at fixed labelled-data budgets.
    Each matrix shows the confusion difference between the pretrained model and
    the from-scratch model at the same budget. Rows denote ground-truth labels and columns denote predicted labels.}
  \Description{Two five-by-five heatmaps shown side by side, one for the 5 percent label
    budget and one for the 100 percent budget. Rows are the ground-truth classes and columns
    the predicted classes, in the order Neutral, Smile, Frown, Eyebrow Raise and Squeezed
    Eyes. Each cell holds the difference between the row-normalized confusion matrix of the
    MAE-pretrained model and that of the from-scratch model at the same budget, encoded with a
    diverging colour scale centred at zero, where positive values favour the pretrained model.
    At the 5 percent budget the diagonal cells for Neutral, Frown, Eyebrow Raise and Squeezed
    Eyes are strongly positive and many off-diagonal cells are negative; the diagonal gain for
    Neutral is larger than the corresponding gain the from-scratch model obtains by going from
    5 to 100 percent labels. At the 100 percent budget the differences are smaller overall and
    are concentrated in the Neutral column, indicating that pretraining mainly reduces
    expressive samples being misclassified as Neutral.}
  \label{fig:confusion_pre_fs}
\end{figure}

\paragraph{Pretraining analysis on individual facial expressions}
We further investigate how MAE pretraining affects class-level predictions at
two labelled-data budgets, 5\% and 100\%. Using the same fold aggregation and
normalization procedure, \autoref{fig:confusion_pre_fs} reports the
element-wise difference
$\mathrm{CM}_{\mathrm{pretrained}}-\mathrm{CM}_{\mathrm{scratch}}$ at each
budget. At the 5\% label budget, the improvement on the Neutral class provided
by pretraining is even larger than the improvement obtained by increasing the
from-scratch model from 5\% to 100\% labelled data. Moreover, MAE pretraining
yields clear classification improvements for Neutral, Frown, Eyebrow Raise, and
Squeezed Eyes, while substantially reducing several off-diagonal confusions.
These results suggest that pretraining is most beneficial when labelled data
are scarce, as it provides a feature representation that already captures much
of the underlying expression structure before supervised FER training. At the
100\% label budget, the gains from pretraining mainly arise from reducing
samples from other expression classes being misclassified as Neutral.

\subsection{Sensor-modality Ablation}
\begin{table}[t]
  \centering
  \caption{Sensor-modality ablation for FER. Navigation and proximity
    channels carry most of the discriminative signal.}
  \label{tab:sensors}
  \small
  \setlength{\tabcolsep}{3pt}
  \begin{tabular}{lcccc}
    \toprule
    Input             & Acc               & F1                & AUC & MCC \\
    \midrule
    $\ast$ Baseline   & $0.200{\pm}0.000$ &
    $0.162{\pm}0.000$ &
    $0.500{\pm}0.000$ &
    $0.000{\pm}0.000$                                                     \\
    IMU
                      & $0.667{\pm}0.006$ & $0.373{\pm}0.009$
                      & $0.786{\pm}0.005$ & $0.343{\pm}0.011$             \\
    Euler
                      & $0.645{\pm}0.006$ & $0.305{\pm}0.006$
                      & $0.730{\pm}0.009$ & $0.254{\pm}0.012$             \\
    Prox
                      & $0.750{\pm}0.015$ & $0.611{\pm}0.017$
                      & $0.885{\pm}0.011$ & $0.544{\pm}0.024$             \\
    Nav
                      & $0.872{\pm}0.009$ & $0.805{\pm}0.013$
                      & $0.965{\pm}0.004$ & $0.771{\pm}0.015$             \\
    IMU + Euler
                      & $0.662{\pm}0.009$ & $0.388{\pm}0.007$
                      & $0.791{\pm}0.004$ & $0.345{\pm}0.011$             \\
    Prox + Nav
                      & $0.888{\pm}0.006$ & $0.833{\pm}0.008$
                      & $0.972{\pm}0.003$ & $0.800{\pm}0.011$             \\
    All
                      & $0.892{\pm}0.005$ & $0.836{\pm}0.007$
                      & $0.970{\pm}0.003$ & $0.808{\pm}0.009$             \\
    \bottomrule
  \end{tabular}
\end{table}

We perform ablation studies to understand which modality impacts FER
performance the most. For each sensor-modality combination in
\autoref{tab:sensors}, we train a separate supervised from-scratch baseline
using only the selected input channels.

We report in \autoref{tab:sensors} the cross-validation performance, in terms
of accuracy, F1-score, AUC, and MCC, across different sensor modalities. A
random baseline is also included for reference. We find that the optical
facial-sensing channels lead to higher performance: navigation is the single
modality with the highest results, and when combining navigation and proximity
sensors we find that the performance is very similar to that obtained when
using all modalities. IMU and Euler-angle features lead to lower performance
when used alone: we speculate that gross head motion and orientation are not
sufficient alone for FER. Nonetheless, using all sensors gives slightly higher
accuracy, F1, and MCC than proximity and navigation alone (which retains a
marginally higher AUC), suggesting that motion and orientation features
potentially provide complementary context in a limited amount of cases.

\section{Discussion}
\label{sec:discussion}

\paragraph{Why label efficiency matters here}
Personalized affect sensing on smart eyewear must adapt to new wearers whose
expression morphology was never seen during training. Gathering many labelled
expressions from every new user is impractical, so the operative question is
not only ``how accurate is the model with all labels'' but ``how little
labelled data does it need to become useful.'' Our results indicate that MAE
pretraining shifts the accuracy-vs-labels curve upward, with the clearest
benefit when labels are scarce.

\paragraph{Generalization across users and tasks}
Both the MAE-pretrained model and the from-scratch baseline remain
substantially above random baseline under participant-disjoint evaluation, even
when trained with only 5\% of the labelled data. This indicates that OMG
signals contain expression-related patterns that generalize to unseen users
rather than relying only on participant-specific characteristics. A notable
finding is that the benefits of SSL pretraining transfer across both tasks and
participant cohorts. The CLUES recordings were not collected for FER, and the
pretraining and downstream datasets contain disjoint participants.
Nevertheless, MAE pretraining on CLUES consistently improves FER on unseen
users. This suggests that pretraining captures reusable facial-sensor dynamics
instead of features specific to a particular task or cohort. In practice,
useful FER representations may therefore be learned from unlabelled recordings
collected during other smart-glass activities, reducing the need to label large
amounts of expression data.

\paragraph{Positioning against supervised state-of-the-art}
As mentioned in \autoref{sec:related_work}, researchers have created
end-to-end fully supervised approaches to perform FER using OCOsense. The
difference in reported FER performance between our approach and their
state-of-the-art results can be attributed to differences in experimental
protocols. For example, we had a different dataset and user pool compared to
\citet{kiprijanovska2023towards}, who trained and validated over 27
participants while we have data from 75 users. Moreover,
\citet{marinova2025deep} performed FER over six classes, i.e., they included an
``other'' class, which we did not use. Indeed, we position our contribution as
orthogonal to existing work, i.e., we focus our work on investigating the
impact of label availability more than creating an end-to-end state-of-the-art
system.

\paragraph{Relationship to personalization}
Prior OCOsense work suggested SSL as a promising approach for addressing
person-specific variation~\cite{kiprijanovska2023towards}. Our
person-independent results provide initial support for this direction: MAE
pretraining improves FER on unseen users relative to an otherwise identical
from-scratch model, with the largest benefit when labelled data are scarce.
Cross-user generalization and personalization are complementary in this
setting. A pretrained encoder can provide a general representation of
facial-sensor dynamics, which could subsequently be adapted to a new wearer
using a small amount of labelled calibration data. It could also support
continual adaptation using unlabelled recordings collected during regular use.
Such adaptation may help account for individual facial morphology, differences
in glasses fit and placement, device re-mounting, and changes across recording
sessions. Although we do not explicitly personalize or adapt models in this
work, future studies should quantify per-user improvements, determine how much
calibration data are required, and evaluate whether continual adaptation
remains stable over extended deployments.

\paragraph{Privacy, ethics, and responsible deployment}
Although OCOsense does not use a camera or microphone, facial-expression
inferences remain sensitive personal data. Deployment therefore requires
informed and revocable consent, clear purpose limitation, and
privacy-preserving processing where feasible. Performance should also be
evaluated across demographic and cultural groups because expression morphology
and display norms vary between users. Finally, predicted expressions should be
treated as behavioral signals rather than direct measurements of emotional
state, particularly in workplace or health-related applications.

\section{Limitations and Future Work}

\paragraph{Acted versus real-life expressions}
The labelled expressions in our training data are \emph{acted}, i.e., posed,
voluntary facial movements performed on cue under a controlled protocol. Acted
expressions are typically more exaggerated and stereotyped than spontaneous
expressions~\cite{7990582} and are often recorded as clearly segmented,
isolated expression episodes. This also makes the FER task comparatively easy
for these sensors (high accuracy even at 5\% labels), which likely compresses
the dynamic range of the label-efficiency effect we observe. How well
label-efficient FER trained on acted expressions translates to real-life,
in-the-wild use therefore remains an open question that must be tested
directly---ideally on spontaneous expressions collected during everyday
activities~\cite{jammot2025egoemotion}, across diverse users and sessions, and
over multiple days to capture device re-mounting and day-to-day variation.

\paragraph{Other limitations}
Our evaluation is limited to a single SSL objective and backbone, namely MAE
pretraining with a ViT encoder. This choice allows us to isolate the effect of
pretraining on label efficiency, but it leaves open whether similar gains would
be obtained with other SSL paradigms, such as contrastive or joint-embedding
predictive objectives, or with alternative architectures, including the
convolutional and recurrent models used by prior supervised FER systems. In
addition, the augmentation magnitudes and per-augmentation application
probabilities were not tuned, and we did not ablate the individual contribution
of each augmentation to downstream FER performance.

\section{Conclusion}
We presented a self-supervised approach to facial expression recognition on OMG
smart glasses, using masked-autoencoder pretraining over the glasses'
facial-sensing channels followed by person-independent finetuning. Across
labelled-data budgets from 5\% to 100\%, pretrained models consistently
outperform from-scratch models, demonstrating substantially improved label
efficiency. Remarkably, the MAE-pretrained model trained with only 5\% of the
labelled data already achieves better performance than the from-scratch model
trained with the entire labelled dataset. Moreover, we also investigate which
sensor-modality combination leads to higher performance in FER. We find that
the proximity and navigation channels alone perform on par with using all
sensor modalities: all sensors give slightly higher accuracy, F1, and MCC,
while proximity and navigation give a slightly higher AUC. These results can
guide smart eyewear system design. Overall, we see self-supervised,
label-efficient FER as a practical building block for next-generation smart
eyewear health and well-being systems.

\begin{acks}
  This work is supported by the Swiss National Science Foundation (SNSF) through the grant PZ00P2\_216405 for the project ``XAI-PAC''.
\end{acks}


\bibliographystyle{ACM-Reference-Format}
\balance
\bibliography{references}

\end{document}